\documentclass[10pt,final,journal]{IEEEtran}
\usepackage{amssymb,amsmath,bm,cite}
\usepackage[dvips]{epsfig}
\usepackage[hang,tight]{subfigure}

\title{Grasping and Stabilizing Tumbling Objects with Rotating-Base Multi-Arm Space Manipulators}
\author{Farhad Aghili ~\thanks{This manuscript is the author-accepted version of an article published in IEEE Transactions on Robotics \cite{Aghili-2026}.}}

\begin{document}

\date{}
\maketitle

\begin{abstract}
This paper presents an innovative guidance and control method for a space manipulator that actively maneuvers its base to hold a tumbling object stationary relative to the base during proximity operations. This simplifies motion planning, reduces collision risk, and enables impact-free capture using single- or multi-arm configurations, with or without predefined grasping interfaces. The proposed approach involves two phases. In the pre-grasping phase, a synchronization control aligns the target’s center of mass (CoM) with the system’s instantaneous center of rotation (ICR), virtually rigidizing the target–servicer system to keep the object stationary for reliable grasping. In the post grasping phase, a control strategy based on Hamiltonian dynamics and optimal control stabilizes the combined system by minimizing time, fuel, or energy, while respecting base torque limits and multiple grasping constraints. For structured targets, applied forces and torques stay within their bounds; for unstructured ones, friction constraints prevent slipping. A case study demonstrates the method’s effectiveness in enabling both the subsequent grasping and de-tumbling phases.
\end{abstract}


\section{Introduction}
\IEEEPARstart{S}{pace} manipulator systems (SMSs), also known as chasers or servicers, are designed to perform on-orbit servicing (OOS) tasks such as satellite maintenance, orbital debris removal, and the assembly and upkeep of orbital infrastructure~\cite{Papadopoulos-Aghili-2021,Aghili-2020a,Dimitrov-Yoshida-2004,Aghili-Parsa-2008b,Aghili-2009a,Thienel-Sanner-2007,Aghili-2010n}.
Typical SMSs consist of a servicing satellite have one or more robotic manipulators, each is equipped with a specialized end-effector or grappling mechanism for capturing and manipulating client satellites or other orbital targets. Topologically, an SMS can be modeled as a kinematic chain with two free-moving ends: the base (i.e., the servicing spacecraft) and the end-effector. Due to the dynamic coupling between the manipulator joints and the motion of both the base and the end-effector, control strategies for SMSs are generally categorized into two main approaches:
(i) free floating arm control, where the base is allowed to drift and rotate passively; and
(ii) coordinated control, which stabilizes the base either partially or fully using internal momentum exchange devices (e.g., reaction wheels or control moment gyroscopes) and, when necessary, thrusters. A third control mode that entails the simultaneous coordinated maneuvering of both ends is also feasible in addition to these traditional control methods. Since the target can remain stationary relative to the base throughout the entire proximity operation, this approach offers several advantages for the robotic capture and manipulation of tumbling objects, thereby enhancing mission safety and operational simplicity. This paper presents a comprehensive investigation into such control strategies for space robotic systems.

The success of forthcoming space exploration and utilization missions relies critically on advances in on orbit robotic manipulation and capture capabilities \cite{Papadopoulos-Aghili-2021,Moghaddam-Chhabra-2021,Aghili-2013}. Capture targets may be cooperative or non cooperative, collaborative or non collaborative, with ground observations indicating that many resident space objects are tumbling, which makes autonomous capture particularly challenging. While robotic manipulators have successfully captured cooperative satellites in orbit \cite{Yoshida-2003,Hirzinger-Landzettel-Brunner-Fisher-Preusche-2004,Aghili-Parsa-2008b}, autonomous capture of tumbling targets remains unachieved due to mission constraints, stringent performance requirements, and technological limitations \cite{Goddard-2010}.
Autonomous robotic capture and manipulation of a tumbling satellite is commonly decomposed into three phases: pre capture, contact, and post capture \cite{Papadopoulos-Aghili-2021,Aghili-2024}. During pre capture, trajectory planning and control aim to intercept the target grapple fixture with minimal relative velocity to limit impact forces \cite{Nenchev-Yoshida-1999,Aghili-2011k}. Numerous optimal and vision based motion planning approaches, including online and adaptive methods for free flying robots, have been proposed for this phase \cite{Wei-Liu-2022,Aghili-2022}. The contact phase spans initial contact to secure grasping, for which force motion and impedance control strategies have been developed to ensure safe interaction \cite{Abiko-Lampariello-Hirzinger-2006,Aghili-2016c,Wang-Shi-2021,Aghili-2008c}, including self learning soft grasp approaches based on variable stiffness \cite{Chu-Wu-2018}. In the post capture phase, the end effector secures the target while the manipulator actively damps and stabilizes residual motion. Control and path planning methods for detumbling and stabilization have been widely studied \cite{Dimitrov-Yoshida-2004,Aghili-2009a,Xu-Wenfu-Meng-2019,Aghili-2019e}. Most existing approaches assume that the servicing manipulator system base attitude must be stabilized, either through coordinated control or reactionless manipulation exploiting the null space of internal momentum devices \cite{Ma-Ma-Shashikanth-2006,Aghili-Parsa-2009,Aghili-2008c,Oki-Nakanishi-Yoshida-2008,Aghili-2011k,Lampariello-Hirzinger-2013,Aghili-2012b}.

Effective autonomous capture of tumbling targets requires tightly integrated planning and control that accommodate operational, system, and safety constraints, including time criticality, actuator limits, and collision avoidance. Prior to proximity operations, the servicing spacecraft uses thrusters to translate the system center of mass (CoM), bringing the target within the manipulator workspace. During proximity operations, the system typically operates in either full free floating or partial free floating mode. In full free floating mode, the system CoM remains stationary and the spacecraft passively responds to manipulator motion, whereas in partial free floating mode the spacecraft attitude is actively regulated using momentum devices to maintain functions such as antenna pointing. While extensive research has addressed planning and control for individual phases, namely pre capture, contact, and post capture, their seamless integration under multiple constraints remains a major challenge. Collision risk is particularly acute during approach and must be considered in both pre and post capture trajectory planning \cite{Hovell-Ulrich-2021,Aghili-2009d}. Moreover, smooth phase transitions are essential due to persistent relative motion after capture, motivating the use of impedance control \cite{Abiko-Lampariello-Hirzinger-2006}. To mitigate these challenges, the servicing robot base can be dynamically controlled to translate and rotate so as to null the relative motion between the tumbling target and the spacecraft base.

This work introduces a novel guidance and control strategy for robotic grasping and manipulation of tumbling space objects, based on the concept of virtual rigidization between the target and the robot base through alignment of the Instantaneous Center of Rotation (ICR) with the Center of Mass (CoM). This strategy eliminates relative velocity between the space robot's base and the target even when the latter exhibits multi-axis tumbling, in contrast to methods that rely on matching the target’s rotation axis and spin rate under pure spin \cite{Aghili-2024}. The post-capture phase further incorporates friction constraints and an optimal control formulation, enabling robotic grasping and detumbling of non-cooperative targets with minimal time, energy, or fuel consumption, depending on the controller’s parameter selection. As illustrated in Fig.~\ref{fig:dualarm_phase1}, a synchronization control scheme aligns the SMS’s ICR with the target’s CoM, causing the combined target–robot system to behave as a single rigid body. This effectively renders the tumbling target stationary relative to the robot base during grasping. During the post-grasping detumbling phase (Fig.~\ref{fig:dualarm_phase2}), the robotic arms remain fixed while coordinated torque commands are applied to the base and joints to dampen the system’s motion in a time-optimal manner, subject to actuation limits and grasping constraints. 
This approach allows the entire robotic operation, both pre- and post-grasping phases, to be treated as grasping a stationary target, thereby minimizing dynamic effects and impact at the moment of capture without the need for impedance control or complex collision avoidance, while also enabling friction-based grasping of tumbling objects using multi-arm robots. The paper is structured as follows: Sections~\ref{sec:Multi-axis} and \ref{sec:Pre-Grasping-Control} present the coordinated planning and control strategy for achieving zero relative velocity during approach and grasping by aligning the ICR and CoM. Section~\ref{sec:post-grasping} formulates the grasping forces and moments in relation to base torques. Section~\ref{sec:optimal-control} describes the optimal control strategy under force and torque constraints, Section~\ref{sec:friction-grasping} addresses friction-based grasping during detumbling, and Section~\ref{sec:validation} presents the validation results.

\section{Dynamics Model}
\subsection{Virtual Rigidization for Achieving Zero Relative Velocity} \label{sec:Multi-axis}

\begin{figure*}
\centering {\includegraphics[clip,width=13cm]{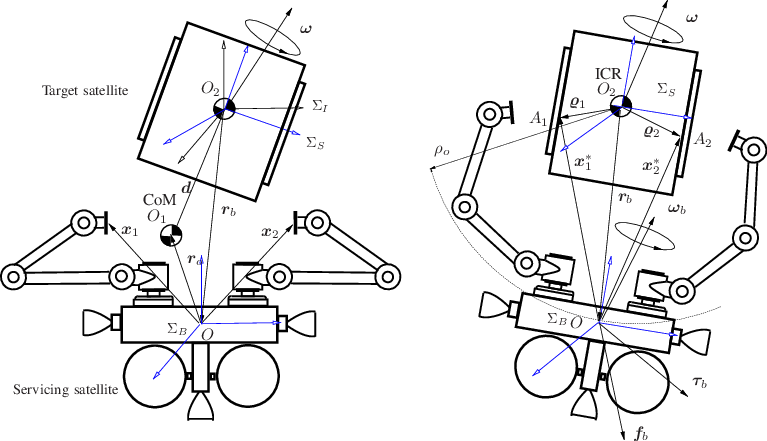}}
\caption{Grasping and stabilizing a tumbling target satellite using a space manipulator system.} \label{fig:dualarm_phase1}
\end{figure*}

Figure~\ref{fig:dualarm_phase1} schematically illustrates a space robotic system attempting to grasp a tumbling space object, where the servicing satellite's base is actively maneuvered to maintain zero relative velocity with respect to the target. To achieve this, the instantaneous center of rotation (ICR) of the servicing satellite must coincide with the target's center of mass (CoM) through coordinated base motion, allowing the combined target-robot system to behave as a single rigid body. This section outlines the dynamic modeling and the design of the coordinated control system.

Suppose the position of the target's centroid is denoted by $O_2$, which coincides with the origin of the body-fixed frame $\Sigma_S$, aligned with the target's principal axes. During the pre-grasping phase, no external forces act on the target, so $O_2$ remains stationary and can be conveniently chosen as the origin of the inertial frame $\Sigma_I$. A separate coordinate frame, $\Sigma_B$, is attached to the base of the space robot at point $O$.
Let ${\bm\omega}$ denote the angular velocity of the target expressed in the body-fixed frame $\Sigma_S$, and let the target's inertia tensor be given by ${\bm I}_c = \mathrm{diag}(I_{xx}, I_{yy}, I_{zz})$, corresponding to its principal moments of inertia. Euler’s rotational equations can then be expressed as:
\begin{equation}
\dot{\bm\omega} = \bm\phi(\bm\omega) \qquad \mbox{where} \quad \bm\phi(\bm\omega)=\begin{bmatrix} \sigma_x \omega_y \omega_z \\ \sigma_y \omega_z \omega_x \\ \sigma_z \omega_x \omega_y \end{bmatrix},
\end{equation}
and vector $\bm\sigma=[\sigma_x\; \sigma_y \; \sigma_z]^T$ contains the following non-dimensional inertia parameters
\begin{equation} \label{eq:p}
\sigma_x = \frac{I_{yy}- I_{zz}}{I_{xx}}, \quad
\sigma_y = \frac{I_{zz}- I_{xx}}{I_{yy}}, \quad
\sigma_z = \frac{I_{xx}- I_{yy}}{I_{zz}}.
\end{equation}
It is worth noting that the inertia ratios are not independent parameters due to the following constraint
\begin{equation} \label{eq:sigma_constraint}
\sigma_x + \sigma_y + \sigma_z + \sigma_x \sigma_y \sigma_z =0.
\end{equation}
Since no external force is applied to the target satellite, the origin of the frame $\Sigma_S$ remains stationary and naturally serves as the instantaneous center of rotation (ICR) prior to capture. The control objective is to apply appropriate forces and torques to the base of the servicing satellite so that its origin also coincides with the ICR, effectively making the two spacecraft behave as if they are rigidly connected. Let ${\bm\rho}$ be a constant displacement vector expressed in the body-fixed frame $\Sigma_S$, representing the position of the servicing satellite’s origin (frame $\Sigma_B$) relative to the target’s origin (frame $\Sigma_S$). Let $\bm{r}_b$ denote the  displacement vector representing the position of the servicing base in the inertial frame. For the point $O_2$ to coincide with the ICR of the base, $\bm{r}_b$ must  be driven to track the following desired displacement
\begin{equation} \label{eq:rb*}
\bm r_b^* = \bm A \bm\rho,
\end{equation}
where $\bm A$ is the rotation matrix describing the orientation of the target, i.e., $\dot{\bm A}= \bm A [\bm\omega \times]$.
Then, the desired velocity and acceleration of the base can be determined as follows:
\begin{equation} \label{eq:v*}
{\bm v}_b^*= \bm A\big( \bm\omega \times \bm\rho  \big),
\end{equation}
where
\begin{equation} \label{eq:dotv*}
\dot{\bm v}_b^*= \bm A\big( \bm\phi(\omega) \times \bm\rho  + \bm\omega \times(\bm\omega \times \bm\rho) \big).
\end{equation}
It is important to note that ${\bm\rho}$ can be interpreted as the length of a virtual rigid link connecting the center of the servicing satellite's base to the target's CoM, which serves as the instantaneous center of rotation for the entire system.

It is worth noting that, although the displacement vector ${\bm\rho}$ can, in principle, be chosen arbitrarily, its selection has practical implications, specifically it influences the velocity that the servicing satellite's base must attain to maintain a stationary relative position with respect to the target. From  \eqref{eq:v*}, we have
\begin{equation} \label{eq:v2}
\| {\bm v}_b^* \| =  \| \bm\omega \|  \bm P {\bm\rho},
\end{equation}
where $\bm P=\bm I - {\bm\omega}{\bm\omega}^T/\| \bm\omega \|^2 =-[\bm\omega \times]^2/ \| \bm\omega \|^2$ is  a projection matrix that projects vectors in $\mathbb{R}^3$ into the plane orthogonal to the angular velocity vector, i.e., $\bm P \bm\omega = \bm 0$.  From equation \eqref{eq:v2}, it is evident that the velocity of the base can be reduced by minimizing the distance $\rho_o=\| \bm\rho \|$.
This distance is constrained by the minimum required separation between the two spacecraft to ensure collision avoidance during proximity operations. Moreover, the expression on the right-hand side of \eqref{eq:v2} suggests that minimizing the base velocity also requires the relative displacement vector to be as closely aligned as possible with the angular velocity vector.

Now we can  express the equation of the  relative angular velocity of the target with respect to the base, in the rotating frame $\Sigma_B$, as follows:
\begin{equation} \label{eq:omega_rel}
\bm \omega_{\rm rel} = \bm\omega_b - \bm A_{\rm rel}(\bm q) \bm\omega,
\end{equation}
where  $\bm\omega_b$ denotes the angular velocity of the base, and $\bm A_{\rm rel}$  representing the relative orientation of frame $\Sigma_S$ relative to frame $\Sigma_B$, corresponding to  the quaternion $\bm q$, i.e.,
\begin{equation} \label{eq:R}
\bm A_{\rm rel}(\bm q) =\bm I + 2 q_o [\bm q_v \times] + 2 [\bm q_v \times]^2,
\end{equation}
where  $\bm q_v$ and $q_o$ are the vector and scalar parts of the
quaternion, $\bm q=[\bm q_v^T \; q_o]^T$. It follows the expression of the relative angular acceleration by
\begin{equation} \label{eq:dot_omega_rel}
\dot{\bm \omega}_{\rm rel} = \dot{\bm\omega}_b - \bm A_{\rm rel} \dot{\bm\omega} - \bm\omega_b \times \bm A_{\rm rel} \bm\omega
\end{equation}

\subsection{Synchronization and Pre-Grasping Control} \label{sec:Pre-Grasping-Control}
In the following analysis, we consider a multi-arm space robot designed to capture a tumbling object. Although only two arms are illustrated in Fig.~\ref{fig:dualarm_phase1}, the formulation is general and applies to an arbitrary number of arms. The index $i$ ($i = 1, 2, \ldots, n$) denotes the variables or parameters associated with the $i$th arm. The dynamic equations of such a space robot  can be expressed in the standard generalized form as follows:
\begin{equation} \label{eq:robot_dynmamics}
\bm M \ddot{\bm\psi} + \bm c(\dot{\bm\psi}, \bm\psi) = \bm\tau - \bm J^T {\cal F}_e.
\end{equation}
Here, $\bm M$ is the generalized inertia matrix of the servicing satellite in the following form ~\cite{Xu-Kanade-1993}
\begin{equation} \label{eq:mass_matrix}
\bm M =\begin{bmatrix} \bm M_{m}  & \bm M_{mb} \\ \bm M_{bm} & \bm M_b \end{bmatrix},  \qquad
\bm M_b=\begin{bmatrix} m_r \bm I  & -m_r [\bm r_o \times] \\ m_r [\bm r_o \times] & \bm M_{b \omega} \end{bmatrix}
\end{equation}
$\bm M_b$ represents the inertia of the entire space robot, $m_r$ is the total mass of the space robot, $[\cdot \times]$ denotes the matrix form of the cross-product, $\bm M_m=\mbox{diag}\{\bm M_{m_1}, \; \bm M_{m_2}\}$ is the inertial matrix of the arms, $\bm M_{mb}=\mbox{diag}\{\bm M_{mb_i}\}$ is the coupled inertia matrix of the base and the arms, and $\bm J=[\bm J_b \; \bm J_m ]$ with $\bm J_b$ and $\bm J_m=[\bm J_{m_1} \;  \cdots \; \bm J_{m_n}]$ being the and-effectors' Jacobians and the arms' Jacobians defined in the based frame $\Sigma_B$. Furthermore, $\dot{\bm\psi}=[{\bm\nu}_b^T \; \dot{\bm\theta}^T ]$ where $\bm\nu_{b}^T=[\bm v_b^T \; \bm\omega_b^T]$ is the generalized velocity of the base of the servicing satellite, vector $\bm\theta^T=[\bm\theta_1^T \; \cdots \;  \bm\theta_n^T]$ contains the joint angles of the arms, $\bm\tau^T=[\bm\tau_m^T \; {\cal F}_b^T]$ is the generalized input force with vector $\bm\tau_{m}^T=[\bm\tau_{m_1}^T\; \cdots \; \bm\tau_{m_n}^T]$ being the torques applied to the joints of the arms, ${\cal F}_e^T=[{\cal F}_{e_1}^T \; \cdots \; {\cal F}_{e_n}^T ]$ is the forces/torques exerted to the end-effectors of the arms expressed in the base frame $\Sigma_B$, ${\cal F}_b^T=[\bm f_b^T \; \bm\tau_b^T ]$ is the generalized force  acting on the space robot base, nonlinear vector $\bm c^T=[\bm c_{m}^T \; \bm c_b^T]$ contains all generalized Coriolis and centrifugal terms, and distance vector $\bm r_o$ connecting $O$ to $O_1$ represents the location of the CoM of the entire space robotic system in the base frame $\Sigma_B$. Note the location of the CoM depends on the mass distribution of the space robot which  changes by changing the joint angles.

Let $m_b$ denote the mass of the servicing satellite base, and let $m_{l_{ij}}$ and $\bm{r}_{{lc}_{ij}}$ represent the mass and the CoM location, respectively, of the $i$th link of the $j$th manipulator, all expressed in the body-fixed frame $\Sigma_B$. Using the definition of the center of mass  for the overall multibody system, the CoM location can then be expressed as a function of the joint angles
\begin{equation} \label{eq:rb}
\bm r_o(\bm\theta) = - \frac{1}{m_b} \sum_{i,j} m_{l_{ij}} \bm r_{{lc}_{ij}}(\bm\theta_j).
\end{equation}
Suppose vector $\bm x^T=[\bm x_1^T \; \cdots \; \bm x_n^T]$ represent the poses of the robotic arms relative to the base, then
\begin{equation}
\dot{\bm x} = \bm J_m \dot{\bm\theta}.
\end{equation}
Now, consider the following coordinated control feedback law for the space robotic system
\begin{subequations}\label{eq:contr_law1}
\begin{equation}
\bm\tau   = \bm c + \bm M \bm u,
\end{equation}
where the auxiliary control input $\bm u^T =[\bm u_m^T \; \bm u_v^T \; \bm u_w^T]$ consists of the following components
\begin{align}
\bm u_m &= K_p \bm J_m^T (\bm x^* - \bm x) - \bm K_d \dot{\bm\theta}\\ \notag
\bm u_v &= \bm A(\bm\phi(\bm\omega) \times \bm\rho + \bm\omega \times (\bm\omega \times \bm\rho))+ \bm K_v (\bm A(\bm\omega \times \bm\rho)-{\bm v}_b) \\ \notag
& + \bm K_r (\bm A \bm\rho - \bm r_b) \\ \notag
\bm u_w &= \bm A_{\rm rel} \bm\phi(\bm\omega) + \bm\omega_b \times \bm A_{\rm rel} \bm\omega  - \bm K_{\omega} \big(\bm\omega_b - \bm A_{\rm rel} \bm\omega \big) - K_q  \bm q_v,
\end{align}
\end{subequations}
$\bm x^*$ denote the desired poses, with all feedback gains being positive definite. Notably, pose selection for minimizing torque and enhancing manipulation capability in robotic grasping was studied in \cite{Ghalamzan-2018}, whereas optimal dual-arm grasp pose selection for redundant space robots was addressed in \cite{Xu-2021}. Given that the servicing satellite does not physically interact with the target during the pre-capture phase, i.e., ${\cal F}_e \equiv \bm 0 $ in  \eqref{eq:robot_dynmamics}, substituting the control law \eqref{eq:contr_law1} into \eqref{eq:robot_dynmamics}, and using the expression for acceleration obtained by differentiating \eqref{eq:omega_rel}, namely, $\dot{\bm \omega}_b = \dot{\bm \omega}_{{\rm rel}} + {\bm \omega}_{{\rm rel}} \times \bm A_{\rm rel} \bm \omega$, yields the closed-loop dynamics of the system as follows:
\begin{align} \label{eq:dot_omeg_dyn}
& \ddot{\bm\theta} + \bm K_d \dot{\bm\theta} + K_p \bm J^T \tilde{\bm x} = \bm 0  \\ \notag
& \ddot{\bm e} + \bm K_v \dot{\bm e} + \bm K_r \bm e  = \bm 0 \\ \notag
& \dot{\bm \omega}_{\rm rel} + \bm K_{\omega} \bm\omega_{\rm rel}  +  K_q  {\bm q}_v  = \bm 0,
\end{align}
where $\tilde{\bm x}=\bm x - \bm x^*$ and $\bm e= \bm r_b - \bm r_b^*$ are the corresponding errors. We define the following positive-definite Lyapunov functions:
\begin{align}
V & = \frac{1}{2} K_p \| \tilde{\bm x} \|^2 +  \frac{1}{2} \| \dot{\bm\theta} \|^2 + \frac{1}{2} \bm e^T \bm K_r \bm e + \frac{1}{2} \| \dot{\bm e} \|^2  \\ \notag
& + \frac{1}{2} K_q \| \bm q - {\bm q}^* \|^2  + \frac{1}{2} \| \bm\omega_{\rm rel} \|^2,
\end{align}
where ${\bm q}^*=[0\;0\;0\;1]^T$ is the quaternion representing the desired relative orientation of the target satellite with respect to the space robot base, and  $\| {\cdot} \|$ denotes the $L_2$ norm or the Euclidean norm of
a vector. Also, consider the following identity, which relates the time derivative of the quaternion to the angular velocity:
\begin{equation}\label{eq:dot_q}
\dot{\bm q} = \frac{1}{2} \bm\Omega(\bm\omega_{\rm rel}) \bm q, \quad \mbox{where} \quad   \bm\Omega(\bm\omega)\triangleq \begin{bmatrix} -[\bm\omega  \times]   & \bm\omega \\
- \bm\omega^T & 0
\end{bmatrix}.
\end{equation}
Then, by substituting \eqref{eq:dot_q} into the time derivative of $V$ long the system trajectories described by \eqref{eq:dot_omeg_dyn}, we obtain
\begin{align} \notag
\dot V = &  - \dot{\bm\theta}^T \bm K_d \dot{\bm\theta} - \dot{\bm e}^T \bm K_v \dot{\bm e} + K_p (\bm q - {\bm q}^* )^T \dot{\bm q} - \bm\omega_{\rm rel}^T  \dot{\bm\omega}_{\rm rel}\\ \notag
= & - \dot{\bm\theta}^T \bm K_d \dot{\bm\theta} - \dot{\bm e}^T \bm K_v \dot{\bm e} + K_p \big( (\bm q - {\bm q}^*)^T \bm\Omega(\bm\omega_{\rm rel}) \bm q \\ \notag
& - \bm\omega_{\rm rel}^T  \Big( \bm\omega_{\rm rel} \times \bm A_{\rm rel} \bm\omega \Big) - K_P  \bm\omega_{\rm rel}^T \bm q_v - \bm\omega_{\rm rel}^T \bm K_{\omega} \bm\omega_{\rm rel} \\  \label{eq:dotV1}
=&- \dot{\bm\theta}^T \bm K_d \dot{\bm\theta} - \dot{\bm e}^T \bm K_v \dot{\bm e}  - \bm\omega_{\rm rel}^T \bm K_{\omega} \bm\omega_{\rm rel} \leq 0,
\end{align}
where the final derivation in the above is concluded using the  following  identities
\begin{subequations} \label{eq:identities}
\begin{equation}
(\bm q - \bm q^*)^T \bm\Omega(\bm\omega_{\rm rel}) \bm q   =  \bm\omega_{\rm rel}^T {\bm q}_v,
\end{equation}
\begin{equation}
\bm\omega_{\rm rel}^T  \Big( \bm\omega_{\rm rel} \times \bm A_{\rm rel} \bm\omega \Big)  = \bm 0.
\end{equation}
\end{subequations}
Therefore, according to the LaSalle's Global Invariant Set Theorem
\cite{Khalil-1992-p115}, the equilibrium point reaches
where $\dot V=0$, or $\bm\omega_{\rm rel} \equiv \bm 0$. Then, by virtue of \eqref{eq:dot_omeg_dyn}  we can infer  identity ${\bm q}_v =0$, which implies  global asymptotic convergence of the orientation error, i.e.,  $\bm\omega_{\rm rel} \rightarrow 0$ and $\bm A_{\rm rel}(\bm q) \rightarrow \bm I$ and $\bm x \rightarrow \bm x^*$ as $t\rightarrow \infty$.

\section{Post-Grasping Optimal Control} \label{sec:post-grasping}

\begin{figure}
\centering {\includegraphics[clip,width=4.5cm]{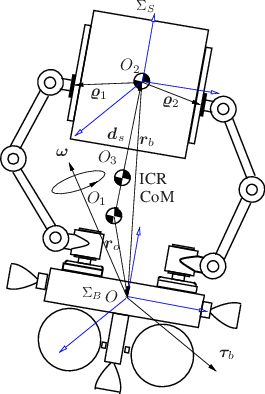}}
\caption{Simultaneous detumbling maneuver of the rigidly interconnected  captured target  satellite and the servicing satellite.} \label{fig:dualarm_phase2}
\end{figure}
\subsection{Modelling of Capture Force and Moment}
The de-tumbling phase begins immediately after the completion of the grasping phase described in the preceding section. To proceed, we derive the relationship between the base actuation torque and the resulting grasping force and moment, which serves as the foundation for the control design presented in the subsequent sections.  As illustrated in Fig.~\ref{fig:dualarm_phase2}, the post-grasping phase begins immediately after the completion of the grasping phase. During this phase, coordinated torque commands are applied to both the base and the manipulator joints to detumble the combined target–robot system as quickly as possible, subject to the torque limits of the servicing satellite’s attitude control system and the grasping constraints associated with the specific grasping configuration. For targets with predefined grasping points, the forces and moments applied to the grapple points are maintained within allowable bounds as will be discussed  Section~\ref{sec:optimal-control}. For targets lacking specific grappling points, friction-based constraints are enforced to ensure contact forces remain within friction cones and only produce pushing actions, thereby preventing slip or slide during the detumbling phase as will be discussed in Section~\ref{sec:friction-grasping}.

In the post-grasping phase, when the target has predefined grasping points, the control objective is to damp out the tumbling motion of the combined servicing and target satellites as quickly as possible, subject to multiple force and torque limitations of the SMS. Specifically, the Euclidean norms of the reaction wheel torque, end-effector torque, and end-effector force must remain within their respective prescribed limits: $\| \bm{\tau}_b \| \leq \alpha_1,\quad \| \bm{\tau}_e \| \leq \alpha_2,\quad \| \bm{f}_e \| \leq \alpha_3$. This is achieved by applying control torques to the reaction wheels, which exert equal and opposite torques on the satellite body. During this detumbling phase, the system states satisfy the following conditions:
$\bm{\omega}_{\rm rel} \equiv \bm{0},\quad \dot{\bm{\omega}}_b \neq \bm{0},\quad \bm{A}_{\rm rel} \equiv \bm{I}$, indicating that the manipulator joints remain stationary while the base alone is responsible for dynamic stabilization. Recall that the base’s axis orientation is aligned with its principal axes.
We now proceed to derive the equations of motion for the combined system by treating the rigidly connected servicing and target satellites as a single rigid body by incorporating the target's inertia into that of the servicing satellite base. Since the manipulator configuration is held fixed during the post-capture phase, i.e., $\dot{\bm{\theta}} = \ddot{\bm{\theta}} = \bm{0}$, the location of the CoM, expressed in the base frame $\Sigma_B$, remains constant. Thus, the vector connecting the CoMs of the target and the servicing satellite is given by
\begin{equation}
\bm d= \bm r_b - \bm r_o.
\end{equation}
Using the parallel axis theorem, the total inertia of the system about the stationary center of mass (CoM) can then be derived as follows:
\begin{equation*}
\bar{\bm M}_{b \omega}  =  \bm M_{b \omega} + \bm I_c + \frac{m_r m_s}{m_r + m_s} \big( \| \bm d\|^2 \bm I - \bm d \bm d^T \big),
\end{equation*}
where $\bm M_{b \omega}$ is one of the sub-matrices of the generalized mass matrix of the space manipulator system \eqref{eq:mass_matrix}. Given that $\bm\omega = \bm\omega_b$ and joint velocities and accelerations are zero, the rotational dynamics of the combined target and SMS simplify to:
\begin{equation} \label{eq:Mbcom-dotomegab}
\bar{\bm M}_{b \omega} \dot{\bm\omega} + \bm\omega \times \bm h = \bm\tau_b,
\end{equation}
where $\bm h$ is the angular momentum of the entire interconnected satellites. Let $m_s$ denote the mass of the target. Then, the displacement vector connecting the target’s CoM, $Q_2$, to the stationary CoM of the entire system, $O_3$, can be expressed as follows:
\begin{equation} \label{eq:ds}
\bm d_s= \frac{m_r}{m_r + m_s} \bm d.
\end{equation}
Thus, the linear acceleration of the target's center of mass can be expressed as:
\begin{equation} \label{eq:ao}
\bm a_o = \dot{\bm\omega} \times \bm d_s + \bm\omega \times(\bm\omega \times \bm d_s).
\end{equation}
By applying the identity in \eqref{eq:ds} in \eqref{eq:ao}, the expression for the generalized force required at the target's CoM to produce its motion during the detumbling operation can be written as
\begin{equation} \label{eq:Fo}
{\cal F}_o = \begin{bmatrix}\frac{m_r m_s}{m_r + m_s}\big( \dot{\bm\omega} \times \bm d + \bm\omega \times(\bm\omega \times \bm d) \big) \\ \bm I_c \dot{\bm\omega} + \bm\omega \times \bm I_c \bm\omega  \end{bmatrix}.
\end{equation}
Finally, by substituting the angular acceleration from \eqref{eq:Mbcom-dotomegab} into \eqref{eq:Fo}, we can establish the relationship between the generalized  force acting on the target and the torque applied to the base of the servicing satellite
\begin{equation} \label{eq:FoGamma}
{\cal F}_o = \bm\Gamma \bm\tau_b + \bm\gamma,
\end{equation}
where $\bm\Gamma^T=[\bm\Gamma_1^T \; \bm\Gamma_2^T]$ and $\bm\gamma^T=[\bm\gamma_1^T \; \bm\gamma_2^T]$ are obtained from
\begin{align} \notag
\bm\Gamma_1 &=  -\frac{m_r m_s}{m_r + m_s} [\bm d \times] \bar{\bm M}_{b \omega}^{-1} \\ \notag
\bm\Gamma_2 &=  \bm I_c \bar{\bm M}_{b \omega}^{-1}  \\ \notag
\bm\gamma_1 &=  -\bm\Gamma_1(\bm\omega \times \bm h) + \frac{m_r m_s}{m_r + m_s} \bm\omega \times(\bm\omega \times \bm d) \\  \label{eq:Gamma12}
\bm\gamma_2 &= - \bm\Gamma_2 (\bm\omega \times \bm h) + \bm\omega \times \bm I_c \bm\omega.
\end{align}
Given the generalized force ${\cal F}_{e_i}^T = [\bm{f}_{e_i}^T \;\; \bm{\tau}_{e_i}^T]$ applied by the end-effector of the $i$th manipulator, the force and torque balance on the target, treated as a rigid body, can be expressed as follows:
\begin{subequations} \label{eq:Fe1}
\begin{align}
\bm T_1 {\cal F}_{e_1} - \bm T_2 {\cal F}_{e_2} & = {\cal F}_{\rm in} \\ \label{eq:TFe=Fo}
\bm T_1 {\cal F}_{e_1} + \bm T_2 {\cal F}_{e_2} & = {\cal F}_o \quad \mbox{where} \quad \bm T_i=\begin{bmatrix} \bm I & \bm 0 \\ [\bm\varrho_i \times] & \bm I \end{bmatrix}
\end{align}
\end{subequations}
is the force/torque transformations from the end-effectors to the target's body frame $\Sigma_S$, $\bm\varrho_i$ denotes the location of the grasping point expressed in $\Sigma_S$, and ${\cal F}_{\rm in}$ represents the internal force. Without loss of generality, we assume zero internal forces for simplicity in the following derivation, i.e.,
\begin{equation} \notag
{\cal F}_{e_2} = \bm T_2^{-1} \bm T_1 {\cal F}_{e_1},
\end{equation}
which in turn can used in  \eqref{eq:TFe=Fo} to  yield
\begin{equation} \label{eq:Fe2Fe1}
{\cal F}_{e_i} = \frac{1}{2} \bm T_i^{-1} {\cal F}_o
\end{equation}
Finally, by substituting the expression for ${\cal F}_o$ from \eqref{eq:FoGamma} into \eqref{eq:Fe2Fe1}, we obtain the following relationship between the end-effector generalized force ${\cal F}_{e_i}=[\bm f_{e_i}^T \;\; \bm \tau_{e_i}^T]^T$ and the actuation torque $\bm{\tau}_b$ applied to the servicing robot's base
\begin{subequations} \label{eq:Fei}
\begin{align}
\bm f_{e_i}&= \frac{1}{2} \big( \bm\Gamma_1 \bm\tau_b + \bm\gamma_1 \big) \\
\bm \tau_{e_i}&= \frac{1}{2} \big( \bm\Gamma_{i+2} \bm\tau_b + \bm\gamma_{i+2} \big)  \qquad i=1,2
\end{align}
\end{subequations}
where
\begin{align} \notag
\bm\Gamma_{i+2} &= -\frac{1}{2}[\bm\varrho_i \times] \bm\Gamma_1 + \bm\Gamma_2  \\ \label{eq:Gammaj2}
\bm\gamma_{i+2} &=  - \frac{1}{2} \bm\varrho_i \times \bm\gamma_1 + \bm\gamma_2 \qquad i=1,2
\end{align}
Similarly, in the following analysis, we establish the relationship between the base torque and the joint torques. Since only torque is applied to the space robot’s base during the detumbling phase, the generalized force at the base simplifies to:
\begin{equation}
{\cal F}_b = \bm N \bm\tau_b,
\end{equation}
where $\bm N=[\bm 0 \; \; \bm I]^T$. Moreover, since the robot joints remain stationary during this phase, i.e., $\ddot{\bm{\theta}} = \bm{0}$, eliminating the base acceleration $\dot{\bm{\nu}_b}$ from \eqref{eq:robot_dynmamics} yields:
\begin{equation} \label{eq:robot_dyn_detumbling}
\bm M_{mb} \bm M_b^{-1}(- \bm c_b + \bm N \bm\tau_b - \bm J_b^T {\cal F}_{e}) + \bm c_{m} = \bm\tau_{m} - \bm J_{m}^T {\cal F}_{e}
\end{equation}
Substituting equations \eqref{eq:Fei} and \eqref{eq:Gammaj2} into \eqref{eq:robot_dyn_detumbling} and rearranging the resulting expression yields
\begin{equation} \label{eq:taumi}
\bm\tau_{m_i} = \bm\Lambda_i \bm\tau_b + \bm\lambda_i \qquad i=1,2
\end{equation}
where
\begin{align} \label{eq:Lamdai}
\bm \Lambda_i &= \bm J_{g_i}^T \bm\Gamma_i + \bm M_{mb_i} \bm M_b^{-1} \bm N \\ \notag
\bm\lambda_i &= \bm c_i - \bm M_{mb_i} \bm M_b^{-1} \bm c_b + \bm J_{g_i}^T \bm\gamma_i
\end{align}
and $\bm J_{g_i}$ is the generalized Jacobian for the $i$th manipulator,
\begin{equation}
\bm J_{g_i}= \bm J_{m_i} - \bm J_b \bm M_b^{-1} \bm M_{bm_i}
\end{equation}
Equations \eqref{eq:Fei} and \eqref{eq:taumi} describe how the attitude control torque applied to the base is transmitted to the end-effector forces/moments and the joint torques of the manipulator arms during the post-capture phase. These relationships are used as constraints in the optimal control design presented in the following section.

\subsection{Optimal De-tumbling Control} \label{sec:optimal-control}
The optimal control problem considered here aims to bring the rigidly connected satellites from an initial angular velocity $\bm{\omega}_b(0)$ to rest while minimizing a specified cost function-such as time, energy, or fuel consumption-subject to actuation constraints. In particular, the Euclidean norms of the reaction wheel torque, end-effector torque, and end-effector force must remain below their respective prescribed bounds $\alpha_1$, $\alpha_2$, and $\alpha_3$. This optimization problem can be formally stated as follows:
\begin{subequations}
\begin{align} \label{eq:one}
\min\quad & \int_{t_0}^{t_f} J \; dt \\
\mbox{subject to:} \quad & \| \bm\tau_b \| \leq \alpha_1 \\
& \| \bm f_{e_i} \| \leq \alpha_2, \quad \| \bm\tau_{e_i} \| \leq \alpha_3\\
& \| \bm\omega(t_f) \| =0.
\end{align}
\end{subequations}
where $J=1$ for time-optimal control, $J=\| \bm\tau_b \|^2$ for  minimum-energy control, $J=\| \bm\tau_b \|^2$ for minimum-fuel control in the case of a single thruster, and $J=| \bm\tau_b |_1=|\tau_{b_x} |+|\tau_{b_y} | +|\tau_{b_z} |$ for minimum-fuel when using discrete thrusters aligned with fixed axes.

Denoting vector $\bm p$ as the costates, we can write the system
Hamiltonian \eqref{eq:Mbcom-dotomegab} and \eqref{eq:one} by
\begin{equation} \label{eq:Hamilton}
H = J + \bm p^T \bar{\bm M}_{b \omega}^{-1} \big(  \bm\omega \times \bm h \big)  + \bm p^T  \bar{\bm M}_{b \omega}^{-1} \bm\tau_b.
\end{equation}
According to optimal control theory \cite{Stengel-1993-p201}, the time derivatives of the costate variables must satisfy
\begin{equation}\label{eq:dot_lambda}
\dot{\bm p}  = -  \frac{\partial H}{\partial \bm\omega} =
-[\bm h \times] \bar{\bm M}_{b \omega}^{-1}  \bm p.
\end{equation}
Substituting the Hamiltonian from \eqref{eq:Hamilton} into \eqref{eq:dot_lambda}, and using the fact that $\partial \bm{h} / \partial \bm{\omega} = \bm{I}_c$, leads to
\begin{equation}
\dot{\bm p} = \bm W \bm p,
\end{equation}
where
\begin{equation} \notag
\bm W= \big( [\bm\omega \times] \bm I_c -[\bm h \times] \big)\bar{\bm M}_{b \omega}^{-1}.
\end{equation}
The solution to the above first-order time-varying differential equation is given by:
\begin{equation} \label{eq:lambdaPhi}
\bm p(t) = \bm\Phi(t, t_0) \bm p_0,
\end{equation}
where $\bm\Phi(t, t_0)$ denotes the state-transition matrix associated with the linear system \eqref{eq:dot_lambda}, and $\bm p_0$ represents the initial value of the costate vector. If $\bm{\tau}_{b}^*$ denotes the time-optimal torque profile, and $\bm{\omega}^*$ and $\bm{p}^*$ are the corresponding solutions of \eqref{eq:Mbcom-dotomegab} and \eqref{eq:dot_lambda} for $\bm{\tau}_{b} = \bm{\tau}_{b}^*$, then, according to Pontryagin's Minimum Principle, the optimal torque $\bm{\tau}_{b}^*$ must satisfy the following condition:
\begin{equation} \label{eq:Pontragin}
\min_{\bm\tau_{b} \in {\cal S}} H(\bm\omega,   \bm p, \bm\tau_{b}),
\end{equation}
where set $\cal S$ represents all admissible control inputs satisfying the constraints, i.e.,
\begin{equation} \label{eq:S}
{\cal S}=\{ \bm\tau_b | \quad  \| \bm\tau_b \| \leq \alpha_1, \; \|\bm f_{e_i} \| \leq \alpha_2,  \; \| \bm\tau_{e_i} \| \leq \alpha_3  \}
\end{equation}
Define directional vector as
\begin{equation}
\bm z = \bar{\bm M}_{b \omega}^{-1} \bm p = \bm Q^T \bm p_0,
\end{equation}
where $\bm Q^T(t)=\bar{\bm M}_{b \omega}^{-1} \bm\Phi(t, t_0)$. Then, combining equations \eqref{eq:Hamilton} and \eqref{eq:Pontragin} implies that the optimal actuation torque vector $\bm{\tau}_b$ should be aligned as closely as possible in the opposite direction of the vector $\bm{z}$, subject to the given constraints. This condition can be equivalently formulated as the following optimization programming:
\begin{subequations} \label{eq:Khun-Tucker}
\begin{align}
\min_{\bm\tau_{b}}  \quad & J + \bm z^T \bm\tau_{b} \\ \label{eq:gg1}
\mbox{subject to:} \quad  & g_j( \bm\tau_{b} )   \leq 0 \qquad j=1,\cdots,4.
\end{align}
\end{subequations}
Here, functions $g_1(\cdot),\cdots, g_4(\cdot)$ are derived  in the following forms from \eqref{eq:FoGamma}, \eqref{eq:Gamma12}, and \eqref{eq:Fe2Fe1}  in conjunction with the force and torque limit constraints given in \eqref{eq:S}
\begin{align}
g_1(\bm\tau_b) =& \bm\tau_b^T \bm\tau_b - \alpha_1^2  \qquad \qquad k=1,2,3 \\ \notag
g_{k+1}(\bm\tau_b) =& \bm\tau_b^T \bm\Gamma_{k}^T  \bm\Gamma_{k} \bm\tau_b
 + 2 \bm\gamma_{k}^T \bm\Gamma_{k} \bm\tau_b + \| \bm\gamma_{k} \|^2 - 4\alpha_{k+1}^2,
\end{align}
where $\alpha_4 = \alpha_3$ is assumed in the above.

Equation \eqref{eq:Khun-Tucker} represents a convex optimization problem in the form of a second-order cone program (SOCP), which can be efficiently solved at each iteration using the Kuhn–Tucker theorem \cite{kuhn-tucker-1951}. However, the constant vector $\bm{p}_0$ remains an unknown that must be determined in the following analysis. By substituting the expression for the optimal actuation torque from \eqref{eq:Khun-Tucker} into \eqref{eq:Mbcom-dotomegab}, the terminal angular velocity can be obtained as a function of $\bm{p}_0$ through integration
\begin{equation} \label{eq:omega_final}
\bm\omega(t_f, \bm p_0) = \int_{t_0}^{t_f} \bar{\bm M}_{b \omega}^{-1} \Big( \bm\tau_b^*(\bm p_0) - \bm\omega \times \bm h  \Big) dt
\end{equation}
where $t_f$ is terminal time. Since the final time $t_f$ is free, the Hamiltonian must vanish at $t_f$, i.e.,
\[  H(t_f) =0. \]
Therefore, using the expression for the Hamiltonian in \eqref{eq:Hamilton}, we arrive at the following additional condition
\begin{align} \label{eq:H=0}
J - {\bm p}_0^T \bm Q(t_f)  \bm\tau_b^*(t_f)=0.
\end{align}
Finally, combining \eqref{eq:omega_final} and \eqref{eq:H=0}, we obtain the following error equation in terms of the unknowns $\{\bm p_0 , t_f\}$
\begin{equation} \label{eq:phi}
\bm\eta(\bm p_0, t_f) =  \begin{bmatrix} \bm\omega(t_f, \bm p_0) \\  J - {\bm p}_0^T \bm Q(t_f)  \bm\tau_b^*(t_f) \end{bmatrix}  = \bm 0.
\end{equation}
The set of four explicit equations in \eqref{eq:phi}, where $\bm{\eta}(\bm{p}_0, t_f) \in \mathbb{R}^4$, can be solved numerically for the unknowns $\{\bm{p}_0, t_f\}$. Clearly, the error evaluated at the correct solution should vanish at the terminal time. Therefore, an unconstrained optimization routine, such as Matlab's \verb"fminunc" function, can be employed to find the solution using a quasi-Newton method \cite{Boyarko-Yakimenko-2011}.

\subsection{Coordination Control for Multi-Arm Grasping} \label{sec:Grasping-Terget}

Once the base torque applied to the space robot is determined, the end-effector forces and moments required to rigidly hold the target using the manipulator arms during the detumbling maneuver can be computed using \eqref{eq:taumi} and \eqref{eq:Lamdai}. To maintain the robot’s desired positions $\bm{x}_i^*$, while ensuring the desired internal force which is zero in this case during the detumbling phase, a leader-follower coordination control strategy is adopted. In this approach, one arm’s position is held firmly either by joint locks or joint-position hold control. Appropriate joint torques are then applied to the other arm to generate the required end-effector force and moment. Without loss of generality, consider a dual-arm system in which the joint positions of Arm 1 are rigidly held. The joint torques for Arm 2 are subsequently determined using the joint torque equation \eqref{eq:taumi}, i.e.,
\begin{equation}
\bm\tau_{m_2} = \bm\Lambda_2 \bm\tau_b + \bm\lambda_2
\end{equation}
The joint positions of Arm 1 can be maintained either by mechanical locks integrated into the joints or by using the following PD controller combined with feedforward disturbance compensation
\begin{equation}
\bm\tau_{m_1} = \bm\Lambda_1 \bm\tau_b + \bm\lambda_1 - K_p \bm J^T_{m_1}(\bm x_1 - \bm x_1^*) -  \bm K_d \dot{\bm\theta}_1.
\end{equation}

\subsection{Friction-based Grasping} \label{sec:friction-grasping}

This section presents a friction-based grasping technique to secure the target during the detumbling phase when predefined grasping interfaces are unavailable. To achieve this, the joint torques of the robotic arms are controlled upon contact with the target to generate sufficient resistance to prevent slipping while minimizing contact forces.

Let $\hat{\bm{n}}_i$ denote the unit normal vector at the $i$th contact point on the target, corresponding to the end-effector of the $i$th arm. The contact force at the tip of the $i$th arm, $\bm{f}_{e_i}$, can be decomposed into normal and tangential components, $\bm{f}_{n_i}$ and $\bm{f}_{t_i}$, respectively. These force components depend on the target’s geometry and can be related to the end-effector force components by:
\begin{equation} \label{eq:fnft}
\bm f_{n_i} = \hat{\bm n}_i \hat{\bm n}_i^T \bm f_{e_i} \quad \mbox{and} \quad  \bm f_{t_i} = (\bm I - \hat{\bm n}_i \hat{\bm n}_i^T) \bm f_{e_i}.
\end{equation}
According to Coulomb’s friction law, the following inequality must be satisfied to prevent slipping or sliding at the contact point
\begin{equation} \label{eq:coulomb}
\| \bm f_{t_i} \| - \mu  \| \bm f_{n_i} \|   <0  \quad \mbox{and} \quad \hat{\bm n}_i^T \bm f_{n_i}<0,
\end{equation}
where $\mu$ is the coefficient of friction at the contact. Alternatively, substituting \eqref{eq:fnft} into inequality \eqref{eq:coulomb} yields the following vector quadratic inequality
\begin{equation} \label{eq:fric_cone}
\bm f_{e_i}^T \bm Z_i \bm f_{e_i} <0  \quad \mbox{and} \quad  \hat{\bm n}_i^T  \bm f_{e_i}<0 \qquad i=1,2, \cdots, n,
\end{equation}
where
\begin{equation}
\bm Z_i= \bm I -(\mu^2 +1 ) \hat{\bm n}_i \hat{\bm n}_i^T.
\end{equation}
In essence, the inequality constraints in \eqref{eq:fric_cone} ensure that the interaction forces stay within the friction cones and apply only pushing, not pulling, forces at the contact surfaces. Moreover, the total contact forces must generate a generalized force consistent with the target’s required motion during detumbling, as determined by the applied base torque and expressed in \eqref{eq:FoGamma}. The resulting force and moment balance can be described by:
\begin{equation} \label{eq:Sfe}
\bm G \bm f_e ={\cal F}_o(\bm\tau_b^*) \quad \mbox{where} \quad \bm G=\begin{bmatrix} \bm I & \cdots & \bm I\\ [\bm\varrho_1 \times ] & \cdots & [\bm\varrho_n \times ] \end{bmatrix},
\end{equation}
where $\bm\varrho_i$  denotes the location of the $i$th contact points, as previously defined. Consequently, to ensure stable grasping during detumbling, both the force/torque balance in \eqref{eq:Sfe} and the friction cone constraints in \eqref{eq:fric_cone} must be satisfied. This leads to a formulation as a quadratically constrained quadratic program (QCQP)
\begin{subequations}
\begin{align} \label{eq:QCQP}
\min_{\bm f_{e} } \quad & \| \bm f_{e} \|^2  \\ \notag
\mbox{subject to:} \quad &  \bm G \bm f_{e}  -  \big( \bm\Gamma \bm\tau_b^* + \bm\gamma \big) = \bm 0 \\ \label{eq:quadratic}
& \bm f_{e_i}^T \bm Z_i \bm f_{e_i} <0 \\ \label{eq:linear}
& \quad \bm n_i^T   \bm f_{e_i} <0 \qquad i=1,\cdots,n
\end{align}
\end{subequations} 
Alternatively, all inequality constraints in \eqref{eq:QCQP} can be reformulated as log-barrier terms within the cost function of the following optimization problem, which is subject only to linear equality constraints
\begin{subequations}
\begin{align}
\min \quad J = &\| \bm f_{e} \|^2 - \beta \log \sum_{j} \varphi_j(\bm f_{e}) \\
\mbox{subject to:} \qquad  & \bm G \bm f_e - {\cal F}_o(\bm\tau_b^*) = \bm 0,
\end{align}
\end{subequations}
where $\varphi_j(\bm f{e})$ includes all quadratic functions from \eqref{eq:quadratic} and linear functions from \eqref{eq:linear}, and $\beta$ is a positive scaling parameter. In summary, the hierarchical control architecture first computes the optimal base torque $\bm{\tau}_b$ from \eqref{eq:Khun-Tucker} and then subsequently uses it in \eqref{eq:QCQP} to determine the required grasping forces. Once the end-effector forces and base torque are obtained from the above optimization program, the joint torques required for grasping can be determined in a similar manner as described in Section~\ref{sec:Grasping-Terget}. To this end, it is worth noting that, in practice, the actual grasping configuration is not precisely known due to measurement errors and control inaccuracies. This issue can be addressed using existing solutions ranging from mechanical compliance to passively accommodate errors, to closed-loop force control with tactile sensing, and learning-based grasp execution methods that inherently compensate for such deviations through trained policies \cite{Kim-Kunihiro-2013,Kim-Jeeseop-2025}.

\subsection{Design Considerations and Integrated Control}

\begin{figure}
\centering {\includegraphics[clip,width=8.5cm]{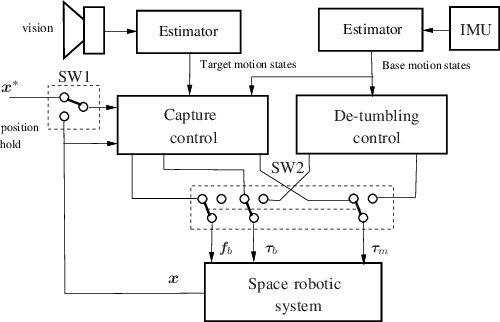}}
\caption{Guidance and control diagram.} \label{fig:control}
\end{figure}

The block diagram in Fig.~\ref{fig:control} illustrates the integrated control architecture and sensory feedback employed during the sequential pre-grasping and post-grasping phases of the capture operation. Throughout the maneuver, the spacecraft base is actively controlled to maintain zero relative velocity with respect to the target. Following base synchronization, the grasping phase begins, during which the manipulator arms move from their initial holding configurations toward the designated grasping points on the target. The post-grasping phase commences once all grasping points have been reached. Sensory feedback may include estimates of the target state and parameters, such as the center-of-mass location, relative position, and angular and linear velocities, obtained from active and passive vision systems operating under the challenging illumination conditions of the space environment \cite{Aghili-Parsa-2009,Lampariello-2021}. The base motion states are estimated using measurements from an onboard inertial measurement unit (IMU). High-precision base torque control can be achieved using either reaction wheels or thrusters, depending on mission requirements such as torque authority, control accuracy, power consumption, mass, and operational lifetime \cite{Markley-2014,Weston-Craig-2025}. The transition from pre-capture to post-capture control is implemented through a simple state-transition logic triggered at the terminal capture time.

It is also worth noting that the manipulator arms must satisfy several key design requirements, including sufficient reach and dexterity to access the target from a range of approach geometries, adequate payload and joint torque capacities to withstand capture-induced loads, and high structural stiffness to ensure precise motion control and stable interaction with the target \cite{Flores-Abad-Ma-2014,Yoshida-Wilcox-2008}.

\section{Demonstration \& Validation Results} \label{sec:validation}

This section presents simulation experiments to demonstrate the end-to-end validation of robotic grasping and detumbling of tumbling objects using the proposed robot control schemes and the concept of velocity matching between the target and servicing satellites while meeting  prescribed functional requirements for the proximity operation \cite{Aghili-Namvar-2008}.

The inertial parameters of the space manipulator system equipped with three identical robotic arms are summarized in Table~\ref{tab:inertia_servicing}. The corresponding initial joint configurations of the three arms are given by  $\bm\theta_1=[1.15\; 1.54 \; 1.60 \; 0.23 \; 0.36 \; 0.37]^T$~(rad), $\bm\theta_2=[1.13\; 1.45 \;1.75 \; 0.22\; 0.32\; 0.28]^T$~(rad), and $\bm\theta_3=[1.12\; 1.41 \; 1.84 \; 0.26 \; 0.32 \; 0.35]^T$~(rad), while the initial  positions or the arm end-effectors are at $[1.1\; 0.3 \; 2.0]^T$, $[-0.8 \; 0.8\; 2.0]^T$, and $[-0.3 \; -1.1  \;2.0]^T$, respectively.  To demonstrate scalability and enable a comprehensive analysis, three case scenarios involving targets with varying mass and inertia properties, as well as distinct initial angular velocities, are considered, as summarized in Table~\ref{tab:inertia_target}.  The target is assumed to have a mass of $m_s = 250$~kg and an inertia matrix $\bm I_c = \mbox{diag}\{70, 100, 150\}$~kg$\cdot$m$^2$ expressed in the target's principal axes coordinate frame $\Sigma_S$ for the first-case scenario. The grasping points on the target are located at $\bm\varrho_1 = [-0.43 \; 0.25 \; 0]^T$m, $\bm\varrho_2 = [0 \; -0.5 \; 0]^T$m, and $\bm\varrho_3 = [0.43 \; 0.25 \; 0]^T$m, expressed in the target's coordinate frame $\Sigma_S$. In this simulation, the base torque actuation is assumed to achieve a precision corresponding to a torque ripple of 0.02\%, while a bias stability of $10^{-3}$deg/hr is assumed for the navigation-grade IMU gyroscope. The actual coefficient of friction at the interface between the target and the robotic arm end-effectors is $0.3$, whereas the controller assumes $\mu = 0.25$ and incorporates a safety factor of 1.2 to account for contact uncertainties \cite{Kim-Kunihiro-2013,Kim-Jeeseop-2025}. We assume that joint friction estimation and compensation during grasping are addressed using established methods, for example as described in \cite{Zhang-2015}. The transition from free motion to contact is modeled using contact dynamics with assumed local contact stiffness and damping  of $10^{6}~\text{N/m}$ and $10^3~\text{N/m/s}$.

\begin{figure}
\centering{\includegraphics[clip,width=8.5cm]{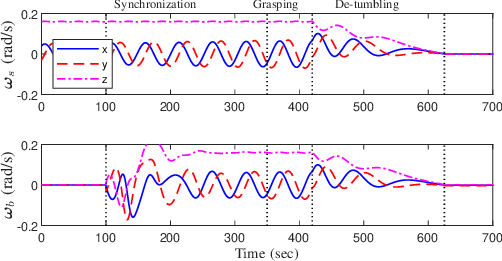}} \caption{The angular velocities of the target satellite and the servicing satellite base.}\label{fig:angular_vel}
\end{figure}

\begin{figure}
\centering{\includegraphics[clip,width=8.5cm]{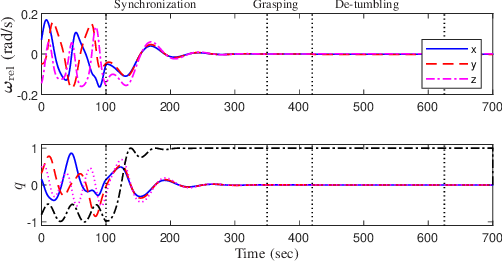}} \caption{The relative angular velocity and orientation of two satellites.}\label{fig:angular_vel_rel}
\end{figure}

\begin{figure}
\centering{\includegraphics[clip,width=8.5cm]{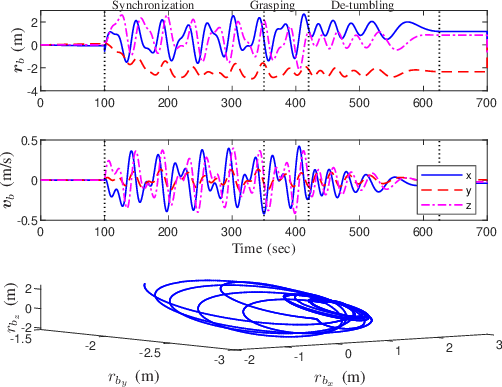}} \caption{Position and linear velocity of the space robot’s base.}\label{fig:base_linear_vel}
\end{figure}

\begin{figure}
\centering{\includegraphics[clip,width=8.5cm]{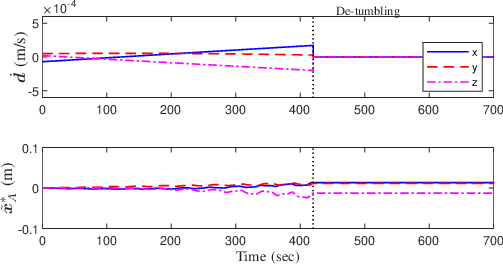}} \caption{Drift velocity and resulting relative position error at point $A_1$.}\label{fig:drift}
\end{figure}

\begin{figure}
\centering{\includegraphics[clip,width=8.5cm]{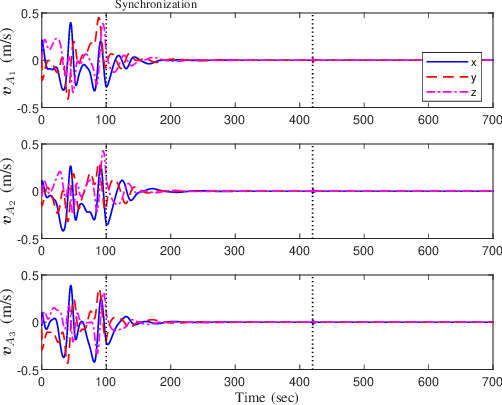}} \caption{The relative velocities of the grasping points $A_1$, $A_2$, and $A_3$ w.r.t. the space robot's base for Case 1.}\label{fig:grasping_vel}
\end{figure}

\begin{figure}
\centering{\includegraphics[clip,width=8.5cm]{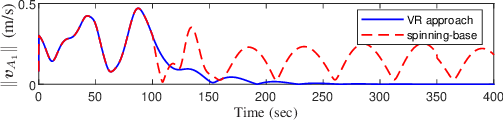}} \caption{Comparison of the relative velocity  at grasping point $A_1$ using the virtual rigidization (VR) method and the base spinning approach.}\label{fig:vel_norm}
\end{figure}

\begin{figure}
\centering{\includegraphics[clip,width=8.5cm]{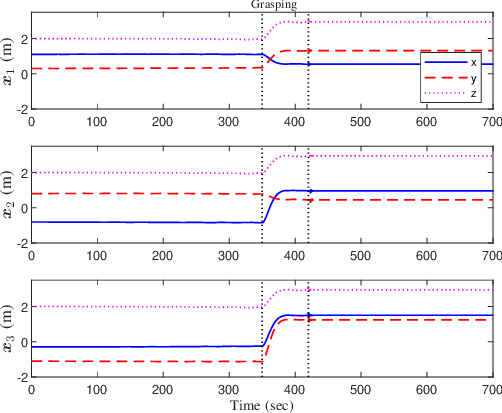}} \caption{Position trajectories of the robotic arms.}\label{fig:ee_position}
\end{figure}

\begin{figure}
\centering{\includegraphics[clip,width=8.5cm]{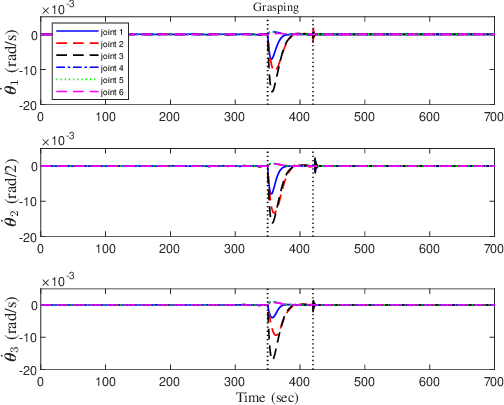}} 
\caption{Joint velocities profiles of the robotic manipulators.}\label{fig:joint_velocity}
\end{figure}

\begin{figure}
\centering{\includegraphics[clip,width=8.5cm]{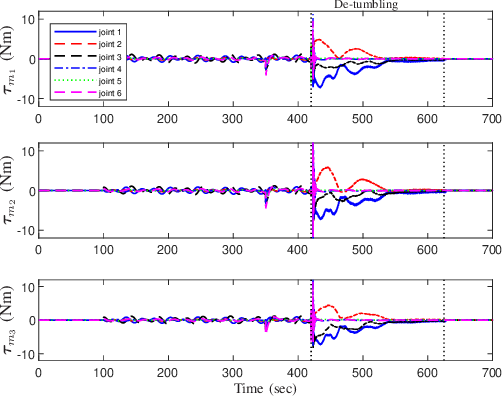}} 
\caption{Joint torque profiles of the robotic manipulators.}\label{fig:joint_torque}
\end{figure}

As shown in Fig.~\ref{fig:angular_vel}, the synchronization controller is activated at $t_0=100$s, when the target has an initial angular speed of $0.167$rad/s. The relative angular velocity and orientation of the target with respect to the robot base are presented in Fig.~\ref{fig:angular_vel_rel}. The position and velocity trajectories of the servicing satellite’s base are plotted in Fig.~\ref{fig:base_linear_vel}, and the velocities of the three grasping points on the tumbling target, expressed in the base frame $\Sigma_B$, are depicted in Fig.~\ref{fig:grasping_vel}.
The plots in Fig.~\ref{fig:drift} illustrate the drift rate between the CoM of the servicing and target satellites, resulting from orbital mechanics effects\footnote{According to Clohessy–Wiltshire equations, the relative acceleration is:
\[ \ddot{\bm d} = \bm L \bm\rho - 2 \bm n \times \dot{\bm d}, \]
where  $\bm L=\mbox{diag}\{3 n^2, 0 , -n^2 \}$ and $\bm n=[0\; 0\;  n]^T$.}, assuming an orbital angular rate of $n=0.001$rad/s and initial measurement drift of $0.1$mm/s. Just before capture, the maximum drift rate reaches $0.22$mm/s. This drift causes the center of mass (CoM) of the servicing satellite to deviate from its initial position on the target’s rotation axis, leading to a gradual increase in misalignment. This drift causes the servicing satellite's CoM to deviate from the target's rotation axis, gradually increasing misalignment and adding to the positioning error of the grasping point relative to the robot's base. As shown in Fig.~\ref{fig:drift},  the resulting error at grasping point $A_1$ reaches approximately $2.6$cm, which remains well within the capture envelope of a typical robotic hand. Fig.~\ref{fig:vel_norm} compares the relative velocity magnitude at grasping point $A_1$ under two scenarios. In the first, the proposed virtual rigidization approach employs coordinated linear and rotational base maneuvers to match the target’s multiaxis angular rate and instantaneous center of rotation. In the second, the base executes a simple spin maneuver to match only the target’s single-axis rotational motion. These results clearly show that synchronization based solely on the base spin rotation cannot eliminate the relative velocity. In contrast, the proposed VR approach, which combines linear and angular base maneuvers, successfully minimizes the relative velocity by $t_1 = 350$s, at which point the target becomes stationary relative to the space robot’s base. This stationary condition substantially facilitates robotic grasping of the target, as evidenced by the end-effector position trajectories shown in Fig.~\ref{fig:ee_position}. The time histories of the joint velocities and joint torques of the multi-arm space manipulator system are shown in Figs.~\ref{fig:joint_velocity} and \ref{fig:joint_torque}, respectively. The results indicate that arm motion occurs primarily during the grasping phase. During the synchronization and de-tumbling phases, the joints remain largely stationary and mainly generate holding or locking torques to maintain the desired joint configurations while counteracting the tumbling motion of the coupled space manipulator system and the forces exerted during grasping.

\begin{figure}
\centering{\includegraphics[clip,width=8.5cm]{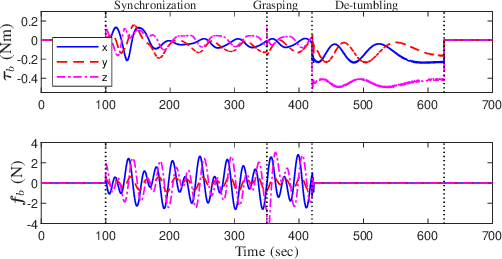}} \caption{Actuation torque and force applied to the space robot’s base.}\label{fig:base_torque_force}
\end{figure}

The optimal torque applied to the robot base, constrained to a maximum base torque of $\alpha_1 = 0.5$Nm, is shown in Fig.~\ref{fig:base_torque_force}, along with the required thrust during the velocity-matching maneuver. The base torque can be generated by reaction wheels or thrusters, while the translational force is produced solely by thrusters. The thrust along each axis can be calculated by multiplying the propellant mass flow rate by the specific impulse ($I_{\rm sp}$):
\[
f_{b_k} = g I_{\rm sp} \dot{m}_p
\]
where $g = 9.81$~m/s$^2$ is Earth's gravitational constant, and $f_{b_k}$ represents the thrust produced by the $k$th thruster. For a typical hydrazine monopropellant thruster with $I_{\rm sp} = 250$s, the estimated propellant mass required to execute the maneuver in Fig.~\ref{fig:base_linear_vel} is 0.25 kg, based on the integration of both positive and negative force components, as shown in Fig.~\ref{fig:base_torque_force}.

\begin{figure}
\centering{\includegraphics[clip,width=8.5cm]{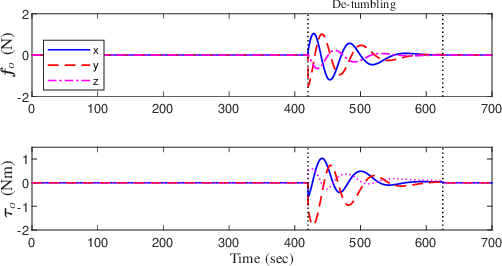}} \caption{Estimation of the inertial forces and moments acting on the target’s centroid during the de-tumbling maneuver.}\label{fig:inerital_force_torque}
\end{figure}

\begin{figure}
\centering{\includegraphics[clip,width=8.5cm]{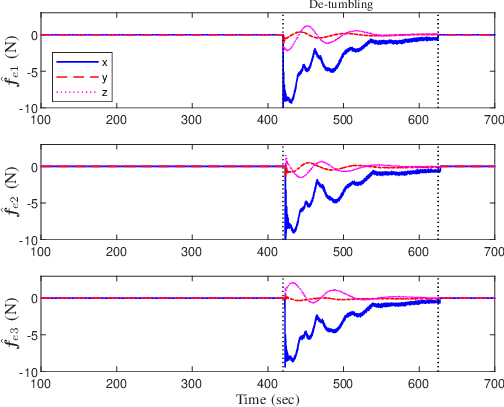}} \caption{The contact forces expressed  at the grasping points.}\label{fig:contact_forces}
\end{figure}

\begin{figure}
\centering{\includegraphics[clip,width=8.5cm]{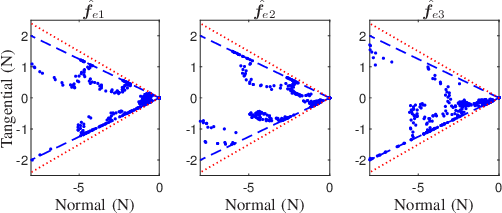}} \caption{Loci of the contact force tangential and normal components w.r.t. the friction cone constraint denoted by hashed-line.}\label{fig:friction_cones}
\end{figure}

The detumbling phase begins at $t_2 = 420$s, following the grasping phase. During this phase, the target, secured by the three robotic arms, remains stationary, while the arms exert forces at the contact points to generate sufficient friction, ensuring no slipping occurs. This is achieved via the prescribed tracking control for detumbling. Fig.~\ref{fig:inerital_force_torque} shows the inertial force and torque exerted on the target’s center during detumbling, while the optimized contact forces at the grasping points are shown in Fig.~\ref{fig:contact_forces}.  The loci of the contact forces shown in Fig.~\ref{fig:friction_cones} confirm that, throughout the detumbling process, the forces remain well within the conservative friction cone defined by $\mu = 0.25$, and are significantly bounded by the actual friction cone corresponding to $\mu = 0.3$ (indicated by the dotted lines). The safety factor 1.2 is selected to enforce uncertainty-aware constraints, ensuring that the planned forces and torques remain valid even under worst-case contact deviations. As shown in Fig.~\ref{fig:base_torque_force}, the base torque is applied from $t_2 = 420$s to $t_3 = 625$s to dampen the angular velocity of the combined system (target and servicing satellite), which are rigidly connected via the robotic arms. Proportional thrusters can generate forces ranging from micro‑Newtons for precision tasks up to several Newtons for orbit control. For example, \cite{Markley-2014,Weston-Craig-2025} report that such thrusters for small satellites span  0.1mN to 3.6N, sufficient to provide the forces required for the simulated maneuvers shown in Fig.~\ref{fig:base_torque_force}. The angular velocity curves in Fig.~\ref{fig:angular_vel} demonstrate that the system comes to rest within $205$s. To this end, recall that the quadratic inequality constraint \eqref{eq:quadratic}, which prevents slipping at the contact points between the target and the robotic arm end-effectors, is equivalent to enforcing that the tangential and normal components of the contact forces remain within the friction cone, as illustrated by the dashed boundaries in Fig.~\ref{fig:friction_cones}. Furthermore, given the contact unit normal vector $\hat{\bm n}_i = [0\; 0 \; 1]^T$, the linear equality constraint \eqref{eq:linear} enforces a negative $z$-component for all contact forces $\bm f_{e_i}$, ensuring purely compressive (pushing) interactions, as clearly reflected in the contact force profiles in Fig.~\ref{fig:contact_forces}.

\begin{figure}
\centering{\includegraphics[clip,width=8.5cm]{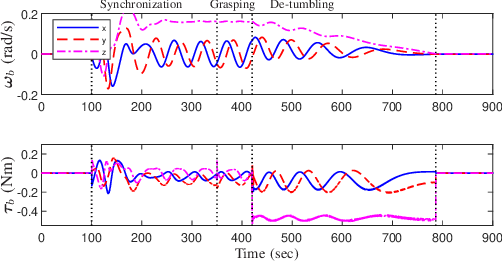}} \caption{Time histories of the base angular velocity and actuation torque for Case 2.}\label{fig:target_case2}
\end{figure}

\begin{figure}
\centering{\includegraphics[clip,width=8.5cm]{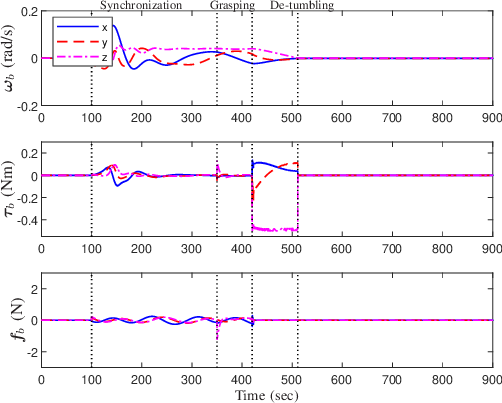}} \caption{Time histories of the base angular velocity, actuation torques, and actuation forces for Case 3.}\label{fig:target_case3}
\end{figure}

Fig.~\ref{fig:target_case2} illustrates the time histories of the base angular velocity and the corresponding actuation torque applied to the base for Case~2. In this scenario, the target is assumed to be approximately 33 times more massive than the space manipulator system, while its initial angular velocity magnitude is maintained at the same level as in Case~1, namely 0.17~rad/s. Consequently, all phases preceding the de-tumbling phase, including the synchronization phase, remain essentially identical to those observed in Case~1. However, as can be inffered from the plots that the system now requires approximately 366~s to complete the de-tumbling process and dissipate the total angular momentum, given the same maximum allowable base torque of $\alpha_1 = 0.5$~Nm. Fig.~\ref{fig:target_case3}  presents the corresponding results for Case~3, in which the target has the same mass and inertia properties as in Case~2, but its initial angular velocity magnitude is reduced to 0.05~rad/s. Consequently, the de-tumbling time is reduced to approximately 91~s under the same torque limit, demonstrating its proportionality to the target's initial angular momentum. As shown in Fig.~\ref{fig:target_case3}, the force required during synchronization to eliminate the relative velocity is significantly reduced due to the lower target angular velocity, and the required propellant mass for executing the maneuver is $0.031$~kg.

\begin{table}
\caption{Inertial parameters of the servicing satellite.}
\begin{center}
\begin{tabular}{ccccc}
\hline \hline
Body & Length  & CoM & Mass   & Inertia \\
     &  (m)  & (m) & (kg)    & (kg-m$^2$) \\
 \hline
Link 2,3   &  2   & 1 & 5  & $ \mbox{diag}[0.01, \; 2, \; 2]$ \\
Link 1,4,5,6   & 0.2 & 0.1  & 1  & $ \mbox{diag}[0.01, \; 0.02, \; 0.02]$ \\
Base & -         & - & 30 & $ \mbox{diag}[10, \; 10, \; 15]$  \\
\hline \hline
\end{tabular}
\end{center}
\label{tab:inertia_servicing}
\end{table}

\begin{table}
\caption{Target inertial parameters and initial angular velocities for the considered scenarios.}
\begin{center}
\begin{tabular}{cccc}
\hline \hline
case &  Mass   & Inertia  & Initial angular rate  \\
   &    kg    & kg-m$^2$ &  rad/s \\
 \hline
1 & 250  & $ \mbox{diag}\{70, 100, 150\}$ & $[0.04 \; -0.03 \; 0.16]^T$ \\
2 & 1000  & $ \mbox{diag}\{280, 400, 600\}$ & $[0.04 \; -0.03 \; 0.16]^T$  \\
3 & 1000  & $ \mbox{diag}\{280, 400, 600\}$ & $[-0.02 \; -0.02 \; 0.04]^T$ \\
\hline \hline
\end{tabular}
\end{center}
\label{tab:inertia_target}
\end{table}

\section{Conclusion}

This paper presented a comprehensive guidance and control framework for robotic capture and stabilization of tumbling orbital objects using a space manipulator system with an actively maneuvering base. The method enabled the target object to remain stationary relative to the servicing satellite during proximity operations, significantly simplifying robotic planning and operational sequences. A sequential strategy was developed, beginning with a synchronization control law that aligned the target’s center of mass with the instantaneous center of rotation of the combined system, virtually rigidizing the target and SMS configuration during the pre-grasping phase. In the post grasping detumbling phase, a control strategy based on Hamiltonian dynamics and optimal control theory was implemented to stabilize the tumbling motion of the combined system by minimizing time, fuel, or energy, subject to base actuation torque limits and grasping constraints. Both structured and unstructured targets were considered, with appropriate force and friction constraints applied to ensure safe and stable interaction during detumbling. A case study demonstrated the effectiveness of the proposed approach under realistic operational conditions, validating the matching base velocity concept for achieving zero relative linear and angular motion. The results confirmed the feasibility of simultaneous grasping and detumbling using a multi-arm robotic system, highlighting the method's potential for future on-orbit servicing missions involving non-cooperative targets.

\bibliographystyle{IEEEtran}


\end{document}